\documentclass{article}
\usepackage{spconf,amsmath,graphicx,hyperref}
\usepackage{amssymb}
\usepackage{graphicx}   
\usepackage{xcolor}    
\usepackage{colortbl}   
\usepackage{array}  
\usepackage{multirow}  
\usepackage{diagbox} 
\usepackage{booktabs} 
\usepackage{caption}   
\usepackage{tabularx}
\definecolor{mygray}{RGB}{230, 230, 230}
\definecolor{bordercolor}{RGB}{215,215,215}
\definecolor{fillcolor}{RGB}{215,215,215}
\usepackage[T1]{fontenc}

\usepackage{graphicx,verbatim}
\usepackage{pifont}
\usepackage{booktabs}

\title{SemaMIL: Semantic-Aware Multiple Instance Learning with Retrieval-Guided State Space Modeling for Whole Slide Images}

\name{
    Lubin Gan\textsuperscript{1$\dagger$}, 
    Xiaoman Wu\textsuperscript{1$\dagger$}, 
    Jing Zhang\textsuperscript{4*}\thanks{$\dagger$ These authors contributed equally. * Corresponding author.}, 
    Zhifeng Wang\textsuperscript{2}, 
    Linhao Qu\textsuperscript{3}, 
    Siying Wu\textsuperscript{4}, 
    Xiaoyan Sun\textsuperscript{1,4*}
}

\address{
    \textsuperscript{1} USTC, Anhui, China,
    \textsuperscript{2} NUDT, Hunan, China,
    \textsuperscript{3} FDU, Shanghai, China, \\
    \textsuperscript{4} Anhui Province Key Laboratory of Biomedical Imaging and Intelligent Processing\\
    Institute of Artificial Intelligence, Hefei Comprehensive National Science Center, Anhui, China \\
}

\begin{document}
\maketitle
\begin{abstract}
Multiple instance learning (MIL) has become the leading approach for extracting discriminative features from whole slide images (WSIs) in computational pathology. Attention-based MIL methods can identify key patches but tend to overlook contextual relationships. Transformer models are able to model interactions but require quadratic computational cost and are prone to overfitting. State space models (SSMs) offer linear complexity, yet shuffling patch order disrupts histological meaning and reduces interpretability. In this work, we introduce SemaMIL, which integrates Semantic Reordering (SR), an adaptive method that clusters and arranges semantically similar patches in sequence through a reversible permutation, with a Semantic-guided Retrieval State Space Module (SRSM) that chooses a representative subset of queries to adjust state space parameters for improved global modeling. Evaluation on four WSI subtype datasets shows that, compared to strong baselines, SemaMIL achieves state-of-the-art accuracy with fewer FLOPs and parameters.
\end{abstract}

\section{Introduction}
\label{sec:intro}

The advent of digital pathology has positioned Whole Slide Images (WSIs) as a pivotal data modality for computational pathology, offering unprecedented opportunities for automated diagnosis and prognosis \cite{gan2025enhancing,wang2025vastsd,wang2024cardiovascular,wang2025angio,li2023ntire,ren2024ninth,wang2025ntire,peng2020cumulative}. Nonetheless, the gigapixel-scale resolution of WSIs, coupled with the scarcity of pixel-level annotations, poses substantial obstacles to the direct application of conventional deep learning techniques. Multiple Instance Learning (MIL)~\cite{gadermayr2024multiple,barbosa2024multiple,amores2013multiple,wang2023decoupling,peng2024lightweight,peng2024towards,wang2023brightness,peng2021ensemble} has therefore emerged as the prevailing paradigm to circumvent these challenges. By obviating the need for exhaustive, fine-grained annotations while still enabling effective exploitation of discriminative cues embedded in large-scale WSIs, MIL furnishes a principled framework that bridges state-of-the-art artificial intelligence methodologies with the stringent requirements of medical image analysis.

\begin{figure}[t]
    \includegraphics[width=\linewidth]{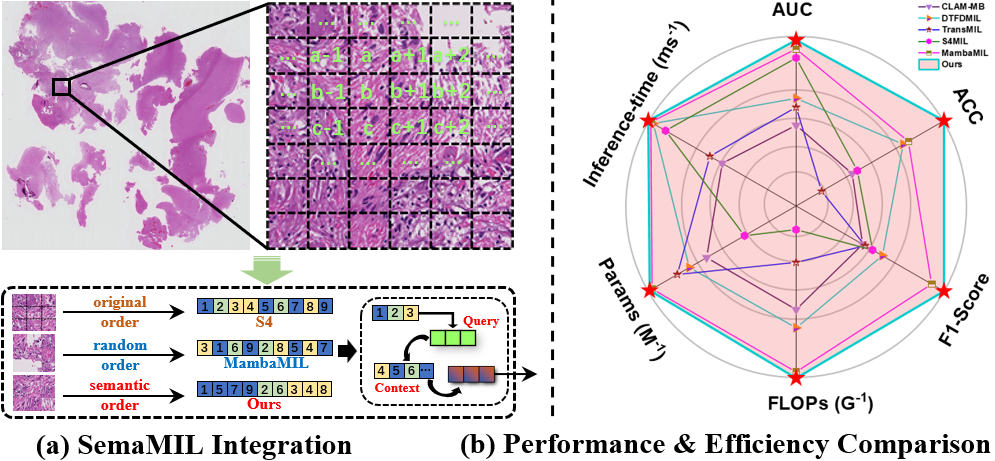}
    \caption{(a) Comparison of patch sequence ordering strategies across different methods for Mamba-based Modeling.  (b) SemaMIL achieves higher computational efficiency and classification performance compared to existing methods.}
    \label{fig:performance}
\end{figure}

Current MIL pipelines typically compress each tissue patch into a low-dimensional embedding with a pretrained encoder~\cite{ming2023towards,ding2024multimodal,chen2024towards,qu2022transmef,qu2022dgmil,qu2022bi,qu2022towards,qu2024rethinking} and then aggregate these embeddings to form a bag-level representation for downstream tasks.     This design turns WSI analysis into a long-sequence modeling problem: a model must capture both the relations among patches and the global context of the whole slide to extract truly discriminative cues.  Attention‐based MIL methods~\cite{ilse2018attention,lu2021data,li2021dual,peng2025boosting,he2024latent,di2025qmambabsr,peng2024unveiling,he2024dual,he2024multi,}, while effective in highlighting discriminative patches, typically treat each patch independently and neglect the contextual dependencies inherent to tissue architecture.    Transformer‐based MIL models~\cite{shao2021transmil,chen2021multimodal,li2021dt,pan2025enhance,wu2025dropout,jiang2024dalpsr,ignatov2025rgb,du2024fc3dnet,jin2024mipi,} can explicitly capture patch interactions, but their self‐attention mechanism incurs quadratic computation and memory costs, leading to prohibitive resource requirements and a propensity to overfit when annotation is scarce.

Recently, state‐space models (SSMs)~\cite{yang2024mambamil,fang2024mammilren2024ultrapixel,yan2025textual,peng2024efficient,conde2024real,peng2025directing,peng2025pixel} have recently emerged as a powerful alternative for long‐sequence modeling.  By providing linear computational complexity alongside a global receptive field, SSMs can efficiently capture long‐range dependencies across thousands of tokens.  These properties render SSMs especially attractive for WSI analysis, where sequences of patch embeddings can easily exceed tens of thousands of elements.

Nevertheless, applying SSMs to MIL directly in pathology encounters its own challenges. Existing pipelines typically randomize the ordering of patch embeddings before sequential processing, thereby discarding histological priors and separating semantically related regions in the sequence \cite{sun2024beyond,qi2025data,feng2025pmq,xia2024s3mamba,pengboosting,suntext}. This arbitrary reordering undermines the model’s ability to exploit tissue‐level context, diminishes interpretability of the learned interactions, and ultimately constrains classification performance~\cite{qu2023rise,qu2022transfuse,qu2023boosting}.

To address these limitations, we propose SemaMIL, a novel Semantic-guided Multiple Instance Learning framework with the following contributions: (1) We introduce a semantic-aware patch ordering mechanism that arranges patches with higher semantic similarity closer together in the sequence, thereby enhancing interaction among histologically relevant regions. (2) We further design a Semantic-guided Retrieval State Space Module (SRSM) to reinforce longrange dependency modeling and data augmentation within the ordered sequence. (3) To evaluate the effectiveness of SemaMIL, we conduct comprehensive experiments on the subtype classification task of whole slide images across four challenging datasets. The results demonstrate that SemaMIL consistently outperforms state-of-the-art methods, further validating its superior performance and robustness.

\begin{figure*}[t]
    \centering
    \includegraphics[width=\textwidth]{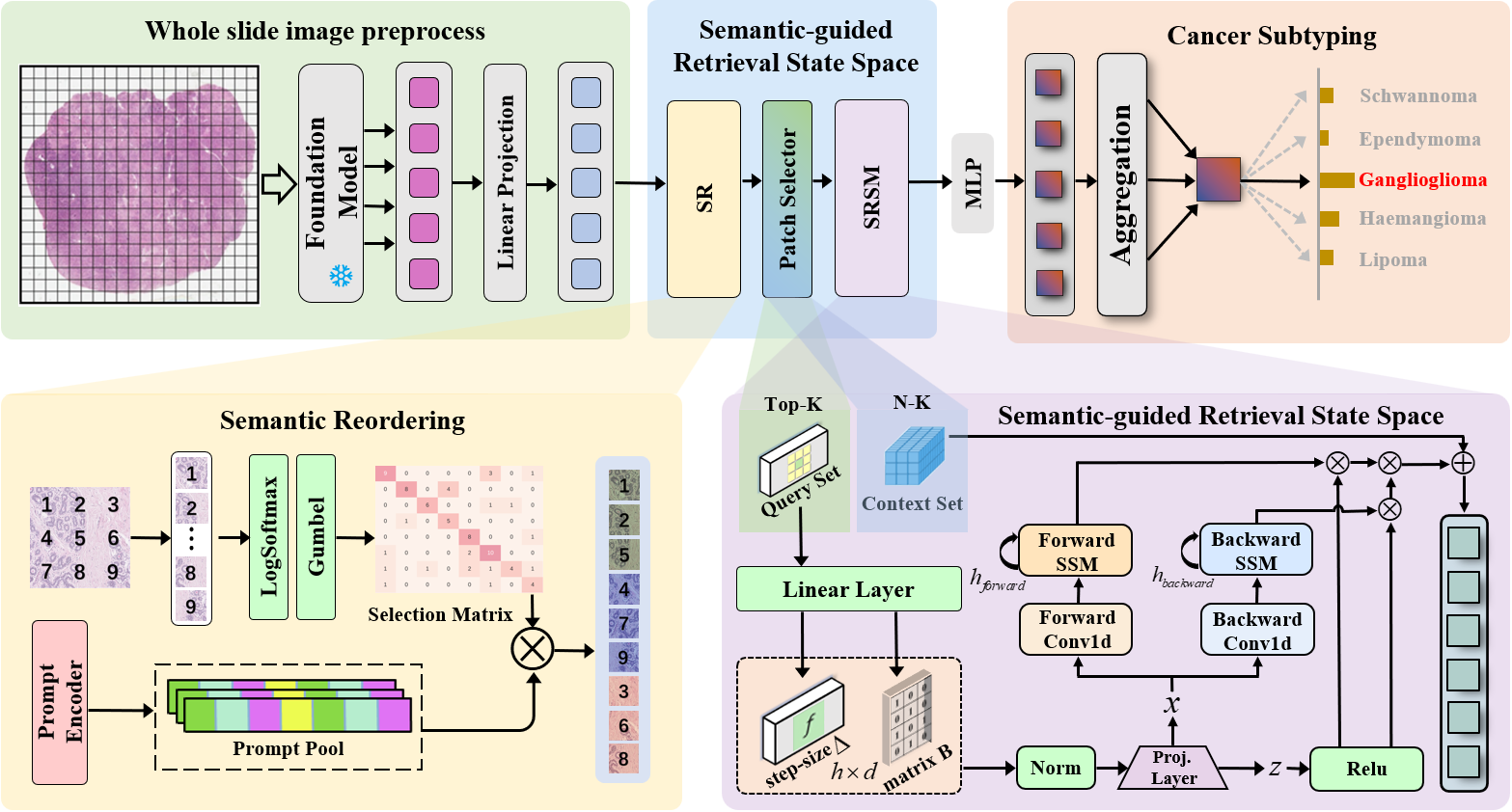}
    \caption{Overview of the SemaMIL framework.}
    \label{fig:mainframe}
\end{figure*}

\section{Methods}
\label{sec:methods}

To better exploit semantic relationships among tissue patches and further enhance long-sequence modeling in multiple instance learning, we propose SemaMIL, which replaces random instance reorderings with a semantically driven rearrangement that brings similar patches closer in the input sequence, and augments state-space sequential modeling with a cross-sequence querying mechanism to reinforce global context.    As illustrated in Fig. \ref{fig:mainframe} , semantic reordering strengthens interactions among related patches, while the augmented framework captures both local and global dependencies under linear computational complexity.

Specifically, given a WSI, we partition its tissue regions into a sequence of $L$ patches $\{ {p_1},{p_2}, \cdots ,{p_L}\}$. A pretrained feature extractor maps these patches to embeddings $X \in \mathbb{R}^{L\times D}$, which are then projected to a lower dimension $d$ by a linear layer. The resulting features undergo the Semantic Reordering module, which computes pairwise similarities and permutes $X$ into a new sequence $X'$ such that semantically similar patches are adjacent. The reordered embeddings $X'$ are passed through a stack of state-space sequential modules enhanced with a global querying operation: each module alternates between scanning the reordered sequence to model instance interactions and querying across all positions to integrate distant contextual information. Finally, an aggregation head pools the refined sequence into a fixed-length bag representation for downstream subtype classification.

\subsection{Semantic Reordering Module}
\label{ssec:semantic reordering module}

In high‐resolution WSI patch sequences, semantically similar but spatially distant patches cannot directly interact within a single causal pass of the state‐space model. To overcome this limitation, we propose a semantic reordering mechanism that adaptively groups related patches in the processing sequence.

As shown in Fig. \ref{fig:mainframe}, given a sequence of patch embeddings $\{ {x_i}\} _{i = 1}^L$ from a WSI, the goal is to place semantically similar patches next to each other so that a single linear state-space pass can more effectively propagate discriminative context. a lightweight router which consists of two linear layers with GELU and produces a semantic score vector for each patch:
\begin{equation}
    {h_i} = GELU({W_1}{x_i}),{z_i} = {W_2}{h_i}.
\end{equation}

We apply a softmax to obtain $p_i$ with $z_i$ and assign a hard semantic label $c_i$ with $argmax(p_i)$, and then form a permutation:
\begin{equation}
    \pi  = \arg sort({c_1}, \cdots ,{c_L})
\end{equation}
to reorder the sequence ${x_{\pi (i)}}$. After processing the reordered sequence with the causal state-space model to obtain outputs ${y_i}'$, we invert the permutation to restore the original spatial order:
\begin{equation}
    y_i = {y_{{\pi ^{ - 1}}(i)}}'.
\end{equation}

This module shortens the effective interaction distance between histologically related regions such as dispersed tumor morphology while avoiding quadratic attention cost. When using soft assignments including Gumbel variants the process is fully differentiable and reversible and provides a semantically compressed input for subsequent retrieval enhanced state space modeling.

\subsection{Semantic-guided Retrieval State Space Module}
\label{ssec:semantic-guided retrieval state space module}

In the aligned and semantically reordered patch sequence $\{ {x_i}\} _{i = 1}^N$, we assign each patch an importance score via a lightweight linear projection and select the top $K$ scoring patches to form the query set $Q = \{ {x_{{i_j}}}\} _{j = 1}^K$, while the remaining $N-K$ patches constitute the context sequence $C = \{ {c_k}\} _{k = 1}^{N - K}$. This Patch Selector effectively suppresses background noise and directs model capacity toward tumor-relevant morphological features. Building on $Q$, we propose SRSM to jointly capture long-range dependencies and global semantic correlations within a unified state-space framework.

We begin with the continuous-time linear time-invariant system:
\begin{equation}
    \begin{aligned}
        \dot{h}(t) &= Ah(t) + Bx(t), \\
        y(t) &= Ch(t) + Dx(t),
    \end{aligned}
\end{equation}
and discretize it via zero-order hold into:
\begin{equation}
    \begin{aligned}
        {h_k} &= {A_d}{h_{k - 1}} + {B_d}{x_k}, \\
        {y_k} &= C{h_k} + D{x_k}.
    \end{aligned}
\end{equation}

To endow the system with dynamic adaptability, we derive its discrete parameters from $Q$ by:

\begin{equation}
  \begin{split}
    \Delta  &= W_\Delta\,\mathrm{vec}(Q) + b_\Delta,\\
    B'      &= W_B\,\mathrm{vec}(Q) + b_B,
  \end{split}
\end{equation}
and then obtain:
\begin{equation}
  \begin{gathered}
    A_d(\Delta)=\exp(\Delta A_0),\\
    B_d(\Delta,B')=(\Delta A_0)^{-1}\bigl(\exp(\Delta A_0)-I\bigr)\,B',
  \end{gathered}
\end{equation}
where $A_0$ is a fixed base transition matrix. The context sequence $\{c_k\}$ is then processed in one causal pass:
\begin{equation}
    \begin{gathered}
        {h_k} = {A_d}(\Delta ){h_{k - 1}} + {B_d}(\Delta ,B'){c_k}, \\
        {y_k} = C{h_k} + D{c_k},
    \end{gathered}
\end{equation}
which $A_0(\Delta)$ captures local tissue continuity, $B_d(\Delta, B')$ gates global semantic querying and noise suppression under the guidance of $Q$. To fully exploit the two-dimensional structure of histopathological slides, we execute this SRSM in parallel along four scan directions, producing outputs $\{ y_k^{(d)}\} _{d = 1}^4$. Finally, global pooling of $y_k^{fused}$ yields a fixed‐dimensional representation, which combines fine‐grained morphological details with holistic semantic context and drives the downstream subtype classification head.

\section{EXPERIMENTS}
\label{sec:exp}

\subsection{Datasets and Evaluation Metrics}

To demonstrate the effectiveness of our proposed SemaMIL, we conduct extensive experiments on a single downstream task, subtype classification of pathology slides using features extracted by the TITAN model~\cite{ding2024multimodal}.   Comparative evaluations are performed on four challenging public datasets: EBRAINS~\cite{roetzer2022digital}, BRACS~\cite{brancati2022bracs}, IPD-Brain~\cite{chauhan2024ipd} and TCGA.

To ensure a robust assessment, we adopt ten-fold Monte Carlo cross-validation and partition each dataset into training, validation and test sets in the ratio of 76.5 percent to 13.5 percent to 10 percent.   For fair comparison with prior work, we also evaluate the official BRACS split indicated by a star in Table \ref{tab:res}. Following standard practice, we report the mean and standard deviation (std) of the area under the ROC curve (AUC) and accuracy (ACC), which together provide a reliable evaluation that mitigates the effects of class imbalance.

\begin{table*}[t]
    \centering
    \caption{Subtype classification results on four datasets. \textbf{Bold}: best performance, \underline{underline}: second-best.}

    \scriptsize
    \renewcommand{\arraystretch}{1} 
    \setlength{\tabcolsep}{2pt} 
    
     \begin{tabularx}{\textwidth}{ c | >{\centering\arraybackslash}X >{\centering\arraybackslash}X | >{\centering\arraybackslash}p{0.8cm} >{\centering\arraybackslash}p{0.8cm} | >{\centering\arraybackslash}X >{\centering\arraybackslash}X | >{\centering\arraybackslash}X >{\centering\arraybackslash}X | >{\centering\arraybackslash}X >{\centering\arraybackslash}X }

    \toprule
    \multirow{2}{1.8cm}{\diagbox{\textbf{Method}}{\textbf{Dataset}}} &  
    \multicolumn{2}{c|}{\textbf{EBRAINS}} & 
    \multicolumn{2}{c|}{\textbf{BRACS$\star$}} & 
    \multicolumn{2}{c|}{\textbf{BRACS}} & 
    \multicolumn{2}{c|}{\textbf{IPD}} & 
    \multicolumn{2}{c}{\textbf{TCGA}} \\

     & AUC & ACC & AUC & ACC & AUC & ACC & AUC & ACC & AUC & ACC \\

    \midrule
    Max-Pooling  & 0.979$\pm$0.005 & 0.714$\pm$0.021 & 0.762 & 0.391 & 0.821$\pm$0.026 & 0.537$\pm$0.055 & 0.861$\pm$0.037 & 0.729$\pm$0.035 & 0.918$\pm$0.019 & 0.733$\pm$0.055 \\
    
    Mean-Pooling & 0.978$\pm$0.003 & 0.714$\pm$0.035 & 0.711 & 0.345 & 0.827$\pm$0.023 & 0.556$\pm$0.045 & 0.880$\pm$0.038 & 0.756$\pm$0.056 & 0.914$\pm$0.013 & 0.699$\pm$0.035 \\
    
    ABMIL~\cite{ilse2018attention}        & 0.967$\pm$0.004 & 0.740$\pm$0.032 & 0.792 & 0.451 & 0.842$\pm$0.021 & 0.575$\pm$0.061 & 0.892$\pm$0.032 & 0.779$\pm$0.046 & 0.924$\pm$0.021 & 0.716$\pm$0.046 \\
    
    CLAM-MB~\cite{lu2021data}      & 0.976$\pm$0.005 & 0.722$\pm$0.033 & 0.796 & 0.461 & 0.836$\pm$0.027 & 0.573$\pm$0.040 & 0.885$\pm$0.037 & 0.760$\pm$0.049 & 0.921$\pm$0.020 & 0.717$\pm$0.045 \\
    
    DSMIL~\cite{li2021dual}        & 0.972$\pm$0.003 & 0.722$\pm$0.031 & 0.787 & 0.452 & \underline{0.856$\pm$0.025} & 0.579$\pm$0.036 & 0.892$\pm$0.032 & 0.769$\pm$0.048 & 0.910$\pm$0.022 & 0.727$\pm$0.034 \\
    
    DTFDMIL~\cite{zhang2022dtfd}      & 0.970$\pm$0.003 & 0.741$\pm$0.023 & 0.802 & 0.462 & 0.853$\pm$0.028 & 0.576$\pm$0.037 & 0.895$\pm$0.036 & \underline{0.781$\pm$0.047} & 0.926$\pm$0.010 & \underline{0.749$\pm$0.037} \\
    
    TransMIL~\cite{shao2021transmil}     & 0.981$\pm$0.003 & 0.711$\pm$0.032 & 0.808 & 0.459 & 0.829$\pm$0.028 & 0.560$\pm$0.055 & 0.860$\pm$0.041 & 0.718$\pm$0.048 & 0.913$\pm$0.018 & 0.689$\pm$0.047 \\
    
    S4MIL~\cite{fillioux2023structured}        & 0.965$\pm$0.003 & 0.715$\pm$0.029 & 0.778 & 0.425 & 0.834$\pm$0.021 & 0.573$\pm$0.036 & 0.897$\pm$0.046 & 0.762$\pm$0.077 & \underline{0.927$\pm$0.019} & 0.735$\pm$0.039 \\
    
    MambaMIL~\cite{yang2024mambamil}     & \underline{0.982$\pm$0.004} & \underline{0.744$\pm$0.034} & \underline{0.820} & \underline{0.467} & 0.855$\pm$0.027 & \underline{0.586$\pm$0.059} & \underline{0.910$\pm$0.034} & \underline{0.781$\pm$0.056} & 0.924$\pm$0.023 & 0.736$\pm$0.048 \\
    
    \textbf{Ours}         & \textbf{0.984$\pm$0.004} & \textbf{0.751$\pm$0.024} & \textbf{0.821} & \textbf{0.472} & \textbf{0.861$\pm$0.025} & \textbf{0.603$\pm$0.036} & \textbf{0.918$\pm$0.025} & \textbf{0.797$\pm$0.047} & \textbf{0.933$\pm$0.023} & \textbf{0.755$\pm$0.047} \\

    \bottomrule
    \end{tabularx}
    \label{tab:res}
\end{table*}

\newcommand{\cmark}{\textcolor{green!70!black}{\ding{51}}}
\newcommand{\xmark}{\textcolor{red}{\ding{55}}}

\begin{table}[t]
    \centering
    \caption{Ablation study for SR and SRSM.}
    \renewcommand\arraystretch{1.05}
    \scriptsize
    \setlength{\tabcolsep}{6pt}
    \begin{tabular}{@{} c c | c c | c c @{}}
        \toprule
        \multicolumn{2}{@{}l|}{\textbf{Our Proposed}}
          & \multicolumn{2}{c|}{EBRAINS}
          & \multicolumn{2}{c}{BRACS} \\
        \cmidrule(lr){3-4}\cmidrule(lr){5-6}
        SR & SRSM & AUC & ACC & AUC & ACC \\
        \midrule
        \xmark & \xmark & 0.969$\pm$0.005 & 0.722$\pm$0.021 & 0.822$\pm$0.028 & 0.561$\pm$0.030 \\
        \xmark & \cmark & 0.977$\pm$0.004 & 0.735$\pm$0.029 & 0.842$\pm$0.026 & 0.578$\pm$0.028 \\
        \cmark & \xmark & 0.979$\pm$0.004 & 0.740$\pm$0.028 & 0.846$\pm$0.025 & 0.584$\pm$0.027 \\
        \cmark & \cmark & 0.984$\pm$0.004 & 0.751$\pm$0.024 & 0.861$\pm$0.025 & 0.603$\pm$0.036 \\
        \bottomrule
    \end{tabular}
    \label{tab:abla1}
\end{table}

\subsection{Implementation Details}

We present the experimental results of our SemaMIL on four datasets compared with the following methods.  These include traditional feature aggregation methods such as Mean Pooling and Max Pooling, ABMIL~\cite{ilse2018attention} and its three variants CLAM-MB~\cite{lu2021data}, DSMIL~\cite{li2021dual}, and DTFDMIL~\cite{zhang2022dtfd}, the Transformer-based TransMIL~\cite{shao2021transmil}, the state-space model-based S4MIL~\cite{fillioux2023structured}, and the recently proposed MambaMIL~\cite{yang2024mambamil}.  In accordance with standard experimental settings, we use the same data pre-processing pipeline as CLAM and apply a fixed learning rate of $5 \times {10^{ - 5}}$ across all methods to ensure optimal performance and enable fair comparisons.

\begin{figure}[t]
  \centering
  \includegraphics[width=\columnwidth]{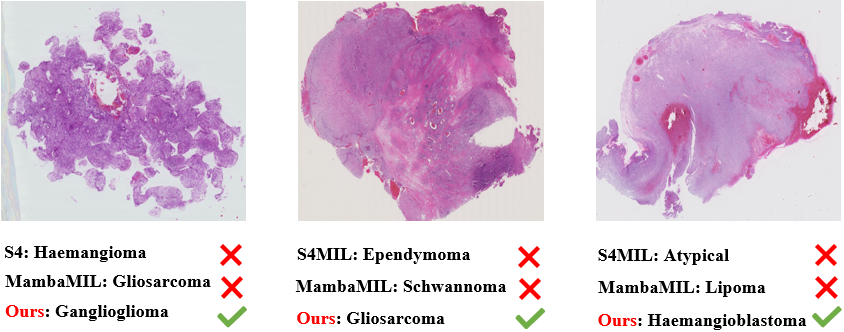}
  \caption{Comparison of classification results among different methods.}
  \label{fig:patch_compare}
\end{figure}

\subsection{Comparison Results}

Table \ref{tab:res} presents the experimental results across four datasets, covering both binary and multi-class classification tasks. Compared to state-of-the-art methods, our proposed SemaMIL consistently surpasses existing approaches in terms of accuracy, achieving an ACC of 75.1\% on EBRAINS and 60.3\% on BRACS, which represent notable improvements over ABMIL and its variants. As illustrated in Fig.~\ref{fig:performance}(a), Mamba-based modeling paradigms adopt distinct patch sequence ordering strategies, and our semantic reordering yields a more coherent arrangement facilitating long-range contextual propagation. Fig.~\ref{fig:performance}(b) shows that our method provides a strong solution to the WSI subtype classification task.

Furthermore, as shown in Table \ref{tab:eff}, SemaMIL demonstrates superior computational efficiency, with FLOPs of 0.2484G and only 0.4641M parameters, both lower than other competing methods. Despite this efficiency, SemaMIL still outperforms other methods in accuracy. Fig.~\ref{fig:patch_compare} presents representative challenging slides misclassified by competing methods but correctly predicted by SemaMIL, serving as illustrative correct cases on difficult subtype instances.

\subsection{Ablation Study}

To assess the effectiveness of SR-Mamba, we conduct extensive experiments to evaluate the contributions of its constituent modules, as shown in Table \ref{tab:abla1}.  The baseline performs a single causal scan in the original spatial patch order.  SR reorders patches according to learned semantic similarity and then applies the causal scan to the reordered sequence.  SRSM preserves the original spatial order while selecting a salient subset of patches as queries to drive state space parameter updates during the full sequence scan.  SR+SRSM applies semantic reordering followed by retrieval driven state space modeling.  As shown in Table 2, these results indicate that both SR and SRSM are necessary and that their combination is the preferred configuration.

\begin{table}[!t]
    \centering
    \caption{Comparison of computational efficiency across different methods.}
    \scriptsize
    \renewcommand{\arraystretch}{1.2}
    \setlength{\tabcolsep}{1pt}
    \begin{tabular}{lccccccccccc}
        \toprule
        \textbf{Methods} & ABMIL & CLAM & DSMIL & DTFDMIL & TransMIL & S4MIL & MambaMIL & \textbf{Ours} \\
        \midrule
        \textbf{FLOPs(G)}      & 0.352 & 0.354 & 0.624 & 0.318 & 0.502 & 0.706 & 0.255 & \textbf{0.248} \\
        \textbf{Params(M)}     & 0.467 & 0.662 & 0.476 & 0.586 & 0.542 & 0.924 & 0.470 & \textbf{0.464} \\
        \textbf{ACC}       & 0.740 & 0.722 & 0.722 & 0.741 & 0.711 & 0.715 & 0.744 & \textbf{0.751}  \\
        \bottomrule
    \end{tabular}
    \label{tab:eff}
\end{table}

\section{Conclusion}

In this work, we introduce SemaMIL, a semantic‑guided multiple instance learning framework for the task of gigapixel whole slide image subtype classification.  Its Semantic Reordering module places histologically related yet spatially distant patches contiguously within a reversible sequence, thereby enhancing information flow in a single causal scan.  The Semantic‑guided Retrieval State Space Module (SRSM) further selects a salient query subset to dynamically modulate state space parameters, reinforcing long‑range dependency modeling, suppressing redundancy, and implicitly enriching contextual interactions with linear time complexity.  Experiments on four challenging datasets demonstrate that SemaMIL consistently benefits from its semantic mechanisms and achieves state‑of‑the‑art performance across all evaluated metrics.  We will further apply our method to a broader range of pathology tasks.

\bibliographystyle{IEEEbib}

\end{document}